\documentclass[sn-mathphys,iicol]{sn-jnl}

\usepackage{graphicx}%
\usepackage{multirow}%
\usepackage{amsmath,amssymb,amsfonts}%
\usepackage{amsthm}%
\usepackage{mathrsfs}%
\usepackage[title]{appendix}%
\usepackage{xcolor}%
\usepackage{textcomp}%
\usepackage{manyfoot}%
\usepackage{booktabs}%
\usepackage{algorithm}%
\graphicspath{{figs/}}
\usepackage{algorithmicx}%
\usepackage{algpseudocode}%
\usepackage{listings}%

\theoremstyle{thmstyleone}%
\theoremstyle{thmstyletwo}%

\theoremstyle{thmstylethree}%

\begin{document}

\title[Article Title]{Data-driven topology design based on principal component analysis for 3D structural design problems}


\author*[1]{\fnm{Jun} \sur{Yang}}\email{yang$\_$jun2023@fuji.waseda.jp}

\author[2]{\fnm{Kentaro} \sur{ Yaji}}

\author[1]{\fnm{Shintaro} \sur{Yamasaki}}

\affil*[1]{\orgdiv{Graduate School of Information, Production and Systems}, \orgname{Waseda University}, \orgaddress{\street{2-7 Hibikino, Wakamatsu, Kitakyushu}, \city{Fukuoka}, \postcode{808-0135}, \country{Japan}}}

\affil[2]{\orgdiv{Graduate School of Engineering}, \orgname{Osaka University}, \orgaddress{\street{2-1 Yamadaoka, Suita}, \city{Osaka}, \postcode{565-0871}, \country{Japan}}}


\abstract{
Topology optimization is a structural design methodology widely utilized to address engineering challenges.
However, sensitivity-based topology optimization methods struggle to solve optimization problems characterized by strong non-linearity.
Leveraging the sensitivity-free nature and high capacity of deep generative models, data-driven topology design (DDTD) methodology is considered an effective solution to this problem.
Despite this, the training effectiveness of deep generative models diminishes when input size exceeds a threshold while maintaining high degrees of freedom is crucial for accurately characterizing complex structures.
To resolve the conflict between the both, we propose DDTD based on principal component analysis (PCA).
Its core idea is to replace the direct training of deep generative models with material distributions by using a principal component score matrix obtained from PCA computation and to obtain the generated material distributions with new features through the restoration process.
We apply the proposed PCA-based DDTD to the problem of minimizing the maximum stress in 3D structural mechanics and demonstrate it can effectively address the current challenges faced by DDTD that fail to handle 3D structural design problems.
Various experiments are conducted to demonstrate the effectiveness and practicability of the proposed PCA-based DDTD.
}


\keywords{Topology optimization, Data-driven topology design, Deep generative model, Principal component analysis}

\noindent



\maketitle

\section{Introduction}\label{sec1}

Topology optimization (TO) aims at designing structures with optimal performance to meet specific engineering requirements and functional demands.
Since TO was proposed by~\citet{bendsoe1988generating}, it has been applied to various engineering problems with tremendous success.
However, with the further development and application of TO methods, some researchers have gradually noticed the difficulty of mainstream sensitivity-based methods in coping with strongly nonlinear problems.
That is, due to the presence of multiple local optima in the solution space of strongly nonlinear problems, sensitivity-based methods may fall into local optima that are low performance in the engineering viewpoint, making it challenging to solve such problems, e.g., turbulent flow channel design~\citep{dilgen2018topology} and compliant mechanism design considering maximum stress~\citep{de2020stress}.
Although researchers have proposed some approaches to improve sensitivity-based methods for dealing with strongly nonlinear problems, these are usually implemented indirectly, potentially leading to issues such as loss of accuracy.
Hence, sensitivity-free methods have gained widespread attention due to their advantages of not requiring sensitivity analysis and offering higher generality.

With the presentation of updating the shape and topology of structures using genetic algorithms by~\citet{hajela1993genetic}, researchers have sequentially proposed a series of sensitivity-free methods~\citep{wang2004graph,tai2005structural,wang2023modified} applied to structural design problems in a given design domain.
Although these methods demonstrate the advantages of sensitivity-free approaches in solving strongly nonlinear problems, they also reveal the difficulty in finding optimal or satisfactory solutions with a high degree of freedom (DOF).
\citet{tai2007target} have attempted to represent material distributions with parametric models to achieve the goal of reducing design variables, but the reduction in the DOF of the design variables also brings the drawback of being difficult to represent complex structures.
Therefore, it is extremely challenging to solve strongly nonlinear problems in a sensitivity-free manner while maintaining the ability to represent material distributions with a high DOF.

With the rapid development of artificial intelligence (AI), some researchers are increasingly recognizing that deep generative models (a type of AI)~\citep{kingma2013auto} have the potential to address the aforementioned problems.
Deep generative models utilize unsupervised learning to extract features from training data and generate novel yet similar data by sampling within the latent space.
Thanks to the capabilities of deep neural networks, deep generative models can generate diverse material distributions with significant flexibility using a limited set of latent variables.

On the basis of this point of view, \citet{yamasaki2021data} proposed a sensitivity-free data-driven topology design (DDTD) methodology for efficiently solving strongly nonlinear multi-objective problems with a high DOF using a deep generative model. 
In their research, elite material distributions are selected from already obtained material distributions with a high DOF on the basis of the non-dominated rank \citep{deb2002fast}.
They are fed into the variational autoencoder (VAE) for training, and their features are extracted into a small-sized latent space.
Then, latent variables are sampled in the latent space and decoded back to the original DOF.
Due to the nature of deep generative models, the newly generated material distributions are diverse and inherit features from the training data, that is, the elite material distributions.
The newly generated material distributions are merged into the training data, and after that, new elite material distributions are selected from the merged data.
They are then used as the inputs for the next round of VAE training. 
By repeating the above processes, the performance of elite material distributions is enhanced while preserving a representation with a high DOF.

Although DDTD is promising to strongly nonlinear structural design problems, it has been realized that limiting the number of DOFs for representing material distributions to a suitable value (approximately tens of thousands in our experience) is crucial for successful VAE training through application studies of DDTD.
However, this limitation significantly impedes the application of DDTD to 3D optimization problems.

On the basis of the above discussions, in this paper, we propose integrating principal component analysis (PCA) into data-driven topology design as a solution to this challenge.
In other words, we train the deep generative model indirectly by utilizing principal component score data obtained via PCA instead of using the original material distribution data.
In this way, the deep generative model generates new principal component score data that inherits the original features while gaining diversity. 
The new material distribution data is subsequently derived by the process of restoration from the new principal component scores.
With the aid of PCA, the DOFs of the material distribution can be reduced from the original DOFs to at most the number of training samples, thereby addressing the challenges faced by DDTD when applied to 3D structural design problems.

In the following, the related work is introduced in section \ref{sec2} and the proposed framework and implementation are described in section \ref{sec3} and its effectiveness is confirmed using numerical examples in section \ref{sec4}. Finally, conclusions are provided in section \ref{sec5}.

\section{Related works}\label{sec2}

\subsection{Topology optimization}
TO has the potential to provide high-performance structural designs that are widely used in industrial manufacturing, 3D printing, medical, and many other fields.
Since the homogenization method of transforming the TO problem for macrostructure into a size optimization problem for the material microstructure was proposed by~\citet{bendsoe1988generating}, TO has gained great focus and motivated more researchers to devote to the TO field.
In order to improve the computational efficiency of TO,~\citet{bendsoe2003topology} proposed the solid isotropic material with penalization (SIMP) method, which performs TO by using the element density as the design variable.
Compared to the homogenization method, the SIMP method can accomplish the TO process more simply, but that is based on sacrificing the physical meaning of the intermediate densities.

In addition, \citet{xie1996evolutionary} proposed the evolutionary structural optimization (ESO) method, i.e., the TO of the structure is achieved by removing the materials in the lower stress region.
In contrast to the above density-based TO methods, researchers have proposed a series of methods for TO of structures by controlling the deformation of the structure via boundary evolution, such as the level-set~\citep{allaire2002level}, moving morphable component (MMC)~\citep{guo2014doing} and moving morphable void (MMV)~\citep{zhang2017explicit}.
The level-set method describes the boundaries of the structures by using the equivalence surfaces of the level-set function and performs TO via the evolution of that function.
The MMC and MMV methods enable topological changes in structures by defining explicit components or holes and controlling their movement and integration.
Other widely developed optimization methods have also brought in fresh ideas on TO domains such as topological derivative~\citep{novotny2012topological} and phase field~\citep{takezawa2010shape}.

It should be noted that this type of sensitivity-based methods build on repeated analysis and design update steps, mostly guided by gradient computation.
However, the reliance on gradient information leads to sensitivity-based methods that may be trapped in local optimal solutions thereby hardly solving strongly nonlinear problems.
Although sensitivity-free TO methods can escape this issue and possess stronger generalization, they also face the challenge of being hardly applicable to TO problems with a high DOF.
Hence, in recent years, part of researchers have expected to utilize machine learning (ML) techniques to solve the above challenges.
Not limited to these challenges, many ML-based TO methods have been proposed as reviewed in the works of~\citet{woldseth2022use,regenwetter2022deep}.
We briefly introduce them as related works in the next section.

\subsection{Machine learning-based TO methods}
In recent years, many works have combined ML with TO to attempt to improve the quality of the solution and reduce the computational cost.
\citet{banga20183d} proposed a deep learning approach based on an encoder-decoder architecture for accelerating TO process. 
\citet{zhang2019deep} introduced a deep convolutional neural network with a strong generalization ability for TO. 
\citet{chandrasekhar2021tounn} demonstrated that one can directly execute TO using neural networks.
The primary concept is to use the network's activation functions to represent the density in a given design domain.
\citet{zhang2021tonr} conducted an in-depth study on the method of directly using neural networks (NN) to carry out TO.
The core idea is reparameterization, which means the update of the design variables in the conventional TO method is transformed into the update of the network's parameters.
\citet{jeong2023physics} proposed a novel TO framework: Physics-Informed Neural Network-based TO (PINN-based TO). 
It employs an energy-based PINN-based TO to replace finite element analysis in the conventional TO to numerically determine the displacement field.

It should be noted that the optimization problems discussed in most of the ML-based TO methods (e.g., minimizing compliance) can be solved as effectively as by using traditional sensitivity-based TO methods.
Although ML-based TO methods have improved in efficiency and effectiveness, they do not address optimization problems that are difficult to solve by sensitivity-based TO methods (e.g., strongly nonlinear problems).

As previously mentioned, the sensitivity-free TO methods can effectively address strong nonlinear problems. 
However, it encounters difficulties when applied to large-scale optimization problems with a high DOF.
With the development and application of ML in TO, some researchers have noticed the potential that exists in deep generative modeling to address this problem.
\citet{guo2018indirect} present a novel approach to TO using an indirect design representation. This method combines a variational autoencoder (VAE) to encode material distributions and a style transfer technique to reduce noise, allowing for efficient exploration of the design space and discovery of optimized structures.
In this research, deep generative models are proved capable of handling large-scale optimization problems by encoding data into latent space.
\citet{oh2019deep} proposed a framework that integrates TO and generative models in an iterative manner to explore new design options, thereby generating a wide array of designs from a limited set of initial design data.
\citet{zhang20193d} explored the 3D shape of a glider for conceptual design and optimization using VAE.
These approaches find the optimal design by employing genetic algorithms (GA) to explore the latent space of the trained VAE. 
Nonetheless, employing randomly generated initial individuals for GA operations results in considerable computational overhead and poses challenges for VAE to learn meaningful features from entirely irregular data.

On the other hand, \citet{yamasaki2021data} proposed a sensitivity-free methodology called DDTD, incorporating a policy to provide the initial material distributions with a certain regularity to ensure that the VAE can effectively capture meaningful features.
In addition, DDTD exclusively utilizes high-performance material distributions to train the VAE, which distinguishes it significantly from another methods that incorporate diverse material distributions as inputs.
With the advantage of sensitivity-free and the capability of solving large-scale problems, DDTD enables to address strongly nonlinear optimization problems that are difficult or even impossible to solve by mainstream TO methods and has been applied in various research fields.
\citet{yaji2022data} proposed data-driven multi-fidelity topology design (MFTD) that enables gradient-free optimization even if tackling a complex optimization problem with a high DOF and applying it to forced convection heat transfer problems.
\citet{kato2023tackling} tackle a bi-objective problem of the exact maximum stress and volume minimization by data-driven MFTD incorporating initial solutions composed of the optimized designs derived by solving the gradient-based TO using the p-norm stress measure. 
\citet{Kii2024latent} proposed a new sampling method in the latent space called the latent crossover for improving the efficiency of DDTD.

\begin{figure*}[t]
\begin {center}
\includegraphics[width=1 \textwidth]{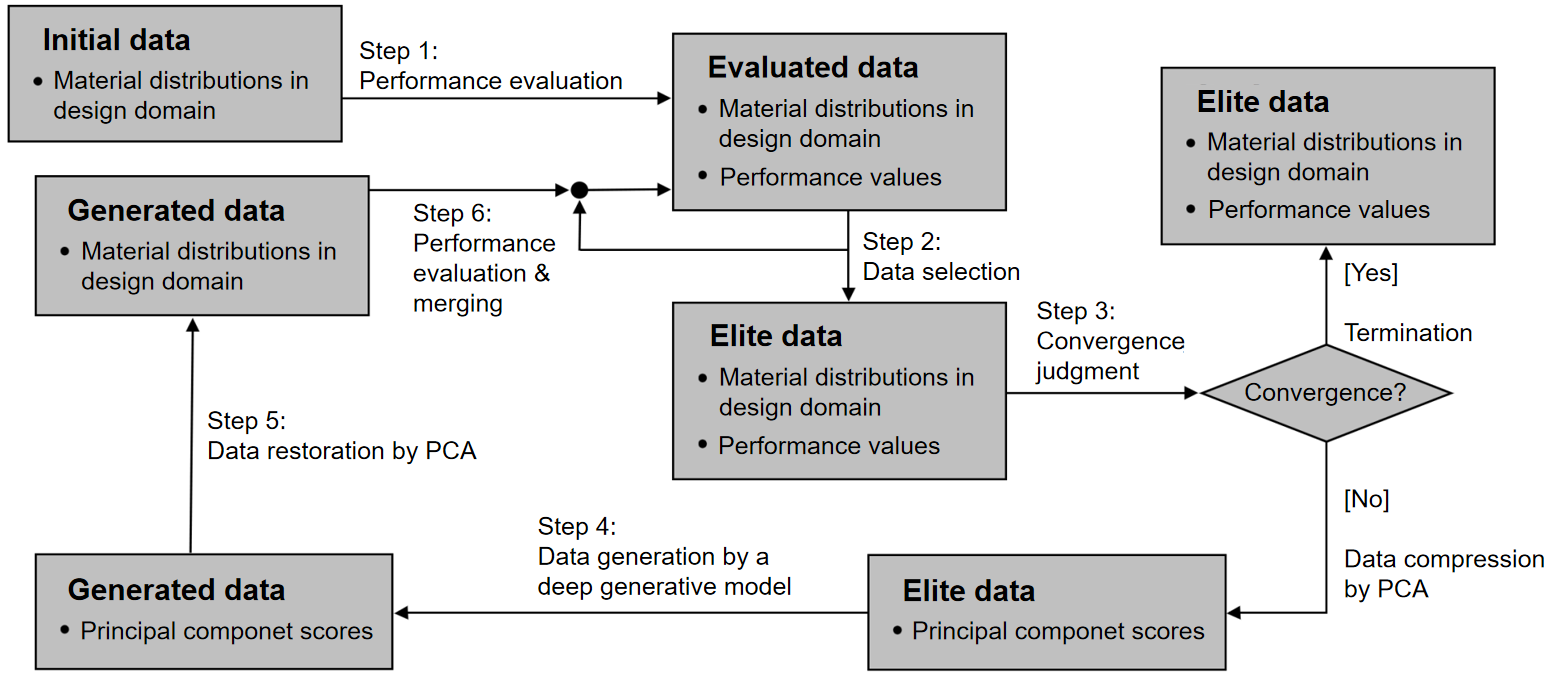}
\caption{Data process flow of the proposed PCA-based DDTD. }
\label{flowchart}
\end {center}
\end{figure*}

\section{Framework and implementation}\label{sec3}

\subsection{Fundamental formulation }\label{sec0}

Consider $D$ as the design domain that is a fixed nonempty and sufficiently regular subset of $\mathbb{R}^{d}(d=3)$ in this paper.
The proposed PCA-based DDTD focuses on the following multi-objective optimization problem in the continuous system:

\begin{equation}
\label{eq1}
\begin{array}{ll}
\underset{\rho}{\text{Minimize}} & \; \left[J_{1}(\rho), J_{2}(\rho), \cdots, J_{N_{\text{obj}}}(\rho) \right] \\ \\
\text{Subject to} & \; G_j(\rho) \leq 0, \;\; \text{for} \; j=1, 2, \dots, N_{\text{cns}}, \\ \\
& \;  0 \leq \rho(\mathbf{x}) \leq 1 .
\end{array}
\end{equation}
Here, $J_{i}$ is the $i$-th objective function and $G_{j}(\rho)$ is the $j$-th constraint function.
The design variable field, the so-called density field, $\rho(\mathbf{x})$ takes values from $0$ to $1$ at an arbitrary point $\mathbf{x}$ in $D$.
$\rho(\mathbf{x}) = 1$ means that the material exists at that point whereas $\rho(\mathbf{x}) = 0$ means the void. $\rho(\mathbf{x})$ has been relaxed according to the manner of the density method.
$N_{\text{obj}}, N_{\text{cns}}$ are the number of the objective and constraint functions, respectively.

For the implementation, the design domain is discretized using a finite element mesh.
On this finite element mesh, the nodal densities serve as the design variables to characterize the corresponding material distribution.
When calculating the objective and constraint functions in Eq.~\ref{eq1}, a body fitting mesh along with the iso-contour of $\rho = 0.5$ is generated for each material distribution.

\subsection{Data process flow }\label{sec31}

\begin{algorithm*}
\caption{PCA-based DDTD }\label{algorithm}
\begin{algorithmic}[1]
\Require Set optimization problem.
\State Build VAE architecture and denoted as $V(*)$
\State Initialize the material distribution data:$\mathbf{X}_{\text{all}} $
\For{$i=0$ to $i_{\text{max}}$}
\State Evaluate performance of $\mathbf{X}_{\text{all}} $: $O_{\text{all}}$ \;
\State Select elite material distribution data: $\mathbf{X}$ $\Leftarrow$ $\mathbf{X}_{\text{all}}, O_{\text{all}}$\;
\If{termination conditions met}
\State break
\EndIf
\State Compress data using PCA: $\mathbf{S}$ $\Leftarrow$ $\mathbf{X}$
\State Generate data via VAE: $\mathbf{S}_{\text{gen}} \Leftarrow V(\mathbf{S})$ \;
\State Restore generated data using PCA: $\mathbf{X_{\text {gen }}} \Leftarrow \mathbf{S}_{\text{gen}}$\;
\State Update material distribution data by merging: $\mathbf{X}_{\text{all}} \Leftarrow {\mathbf{X}}, \mathbf{X_{\text {gen }}}$\;
\State $i \Leftarrow i+1$
\EndFor
\State Obtain satisfactory elite material distribution data: $\mathbf{X}$\;
\end{algorithmic}
\end{algorithm*}

As a sensitivity-free methodology, DDTD uses a deep generative model to generate diverse data that differs from the input data.
However, the capacity of the deep generative model (VAE in this paper) is not infinite. 
This necessitates limiting the DOFs of the input data to a certain value (approximately tens of thousands in our experience) to ensure effective training of the VAE.
Meanwhile, the representation of the designed structure using material distributions with a high DOF is important for characterizing shape and morphological changes, particularly in 3D structural design problems.
In order to address the conflict between both, the data dimensionality reduction method is employed to preprocess the input dataset of VAE.
By using PCA (a data dimensionality reduction method) and DDTD methodology, material distributions with higher performance under the 3D optimization problem are iteratively selected while maintaining their high DOF representation.
Fig.~\ref{flowchart} shows the data process flow of the proposed PCA-based DDTD and the details of each step are explained here.

\textbf{Initial data generation }
Since deep generative models typically struggle to extract meaningful features from highly irregular material distributions, it is essential for the initial dataset to exhibit a certain degree of regularity.
In this paper, we construct a pseudo-problem (low-fidelity problem) that is easily and directly solvable, yet relevant to the original multi-objective optimization problem (high-fidelity problem). 
The solution to this low-fidelity problem is used as the initial dataset.
For example, if the high-fidelity problem is a compliant mechanism design problem (an example is in Sec~\ref{exam2}) that considers geometric nonlinearity, we solve a compliant mechanism design problem under the assumption of linear strain as the low-fidelity problem.
It is important to note that although all numerical examples in this paper utilize material distributions obtained by solving low-fidelity problems, DDTD is not restricted to this approach.
Initial material distributions can also be derived from a parametric model including random numbers, for example. 
Therefore, multi-fidelity problems are not mandatory for DDTD, including PCA-based DDTD.

\textbf{Performance evaluation and data selection} 
We evaluate the performance of the material distributions on the basis of the high-fidelity multi-objective problem, thereby obtaining the evaluated data. 
Subsequently, we perform the non-dominated sorting~\citep{deb2002fast} to obtain the rank-one material distributions as the elite data.

\textbf{Data compression by PCA } 
If the given convergence criterion is satisfied, we terminate the calculation and obtain the current elite data as the final results. 
Otherwise, we conduct data compression by PCA.
Here, we denote the material distributions of the elite data, $\mathbf{X} \in\mathbb{R}^{m \times n}$, as follows:
\begin{equation}
\mathbf{X} = \left[ \begin{array}{c} \hat{\boldsymbol{\rho}}_{1} \\ \vdots \\ \hat{\boldsymbol{\rho}}_{i} \\ \vdots \\ \hat{\boldsymbol{\rho}}_{m} \end{array} \right] ,
\end{equation}
where $\hat{\boldsymbol{\rho}}_{i} \in\mathbb{R}^{1 \times n}$ is the nodal density vector of the $i$-th material distribution, $m$ is the number of the material distributions, and $n$ is the number of DOF for each material distribution.

By using PCA, $\mathbf{X}$ is processed as follows:
\begin{equation}
\label{eq4}
 \mathbf{\bar{X}} =\mathbf{S C}^{\top},
\end{equation}
where $\mathbf{\bar{X}}$ is $\mathbf{X}$ after the centering, $\mathbf{C} \in\mathbb{R}^{n\times m}$ is the principal component coefficient matrix, and $\mathbf{S} \in\mathbb{R}^{m \times m}$ is the principal component score matrix.
Here, it is important that $m$ is independent from $n$.
Therefore, if we limit $m$ to hundreds or thousands (400 in this paper), we can resolve the issue of the input and output size of the VAE by feeding $\mathbf{S}$ into the VAE as the training data.

\textbf{Data generation }
Further, we train the VAE using the principal component score matrix $\mathbf{S}$ instead of the original material distribution $\mathbf{X}$. 
We denote the newly generated principal component score matrix as $\mathbf{S}_{\text{gen}}$.
The principal component scores in $\mathbf{S}_{\text{gen}}$ are diverse and inherit features from those in $\mathbf{S}$ because of the generation process of the VAE.
The architecture of the VAE used in this method is presented in Sec~\ref{VAE}.

\textbf{ Data restoration by PCA } 
Then, we restore material distributions from $\mathbf{S}_{\text{gen}}$.
First, we calculate $\bar{\mathbf{X}}_{\text {gen }} $ using the following equation:
\begin{equation}
\label{eq10}
\bar{\mathbf{X}}_{\text {gen }} =\mathbf{S}_{\text{gen}} \mathbf{C}^{\top} .
\end{equation}

Next, we conduct the inverse operation of the centering to $\bar{\mathbf{X}}_{\text {gen }}$. 
By doing so, we obtain the generated data $\mathbf{X}_{\text {gen }}$.
After that, $\mathbf{X}_{\text {gen }}$ is normalized such that the material distributions have clear and fixed-width transition zones between the solid and void phases.
The details of the normalization are presented in Sec~\ref{nor}.

\textbf{Performance evaluation and merging }
The performances of the generated data are evaluated on the basis of the high-fidelity multi-objective problem. 
Subsequently, the generated data, along with their performance values, are merged into the current elite data. 
In this manner, we update the evaluated data.

Through the above steps, we solve the challenge of applying DDTD to 3D structural design by addressing the conflict between the DOFs of the material distribution and the input constraints of the deep generative model.
To conclude the overview, we summarize the entire procedure in Algorithm~\ref{algorithm}.

\subsection{Normalization}\label{nor}
As described in Sec~\ref{sec0} and Sec~\ref{sec31}, we represent material distributions using the nodal density vector $\hat{\boldsymbol{\rho}}$.
Each component of $\hat{\boldsymbol{\rho}}$ has to take values from 0 to 1 according to the basic concept of TO. 
In addition, it should be 0 or 1 except for the boundaries between the solid and void phases. However, these are not guaranteed on the material distributions generated by the VAE. 
Therefore, we conduct the normalization to those generated material distributions as described below.

First, we give the nodal level-set function $\hat{\boldsymbol{\phi}}$ corresponding to $\hat{\boldsymbol{\rho}}$ generated by the VAE, as follows:
\begin{equation}
\hat{\phi}_i = 2 \hat{\rho}_i - 1 , \;\; \text{for} \; i=1, 2, \dots, n,
\end{equation}
where $\hat{\phi}_i$ and $\hat{\rho}_i$ are the $i$-th components of $\hat{\boldsymbol{\phi}}$ and $\hat{\boldsymbol{\rho}}$, respectively.
Next, we re-initialize $\hat{\boldsymbol{\phi}}$ as the signed distance function, using a geometry-based reinitialization scheme \citep{yamasaki2010structural}.
Finally, the nodal density vector is updated using the following equation:

\begin{eqnarray}
& \hat{\rho}_{\text{u},i} = \left\{\begin{array}{cl}
0 & (\hat{\phi}_i < -h) \\ \\
H(\hat{\phi}_i) & (-h \leq \hat{\phi}_i \leq h) \\ \\
1 & (h<\hat{\phi}_i)
\end{array}\right. , \nonumber
\\
& \qquad\qquad\qquad\qquad \text{for} \; i=1, 2, \dots, n,
\end{eqnarray}
where $\hat{\rho}_{\text{u},i}$ is the $i$-th components of the nodal density vector after the update, $h$ is the parameter for the bandwidth of the transition zone between the solid and void phases, and $H(\hat{\phi}_i)$ is defined as follows:
\begin{equation}
H(\hat{\phi}_i)=\frac{1}{2}+\frac{15}{16}\left(\frac{\hat{\phi}_i}{h}\right)-\frac{5}{8}\left(\frac{\hat{\phi}_i}{h}\right)^{3}+\frac{3}{16}\left(\frac{\hat{\phi}_i}{h}\right)^{5}.
\end{equation}
By the above calculation, each component of the nodal density vector after the update becomes 0 or 1 except for the transition zone between the solid and void phases. 
In addition, the bandwidth of the transition zone is fixed with $2h$.
In the numerical examples of this paper, we set $h$ as 0.01, which is the element length of the finite element mesh discretizing the design domain, if we describe nothing.

\begin{figure}[t]
\begin {center}
\includegraphics[width=0.48 \textwidth]{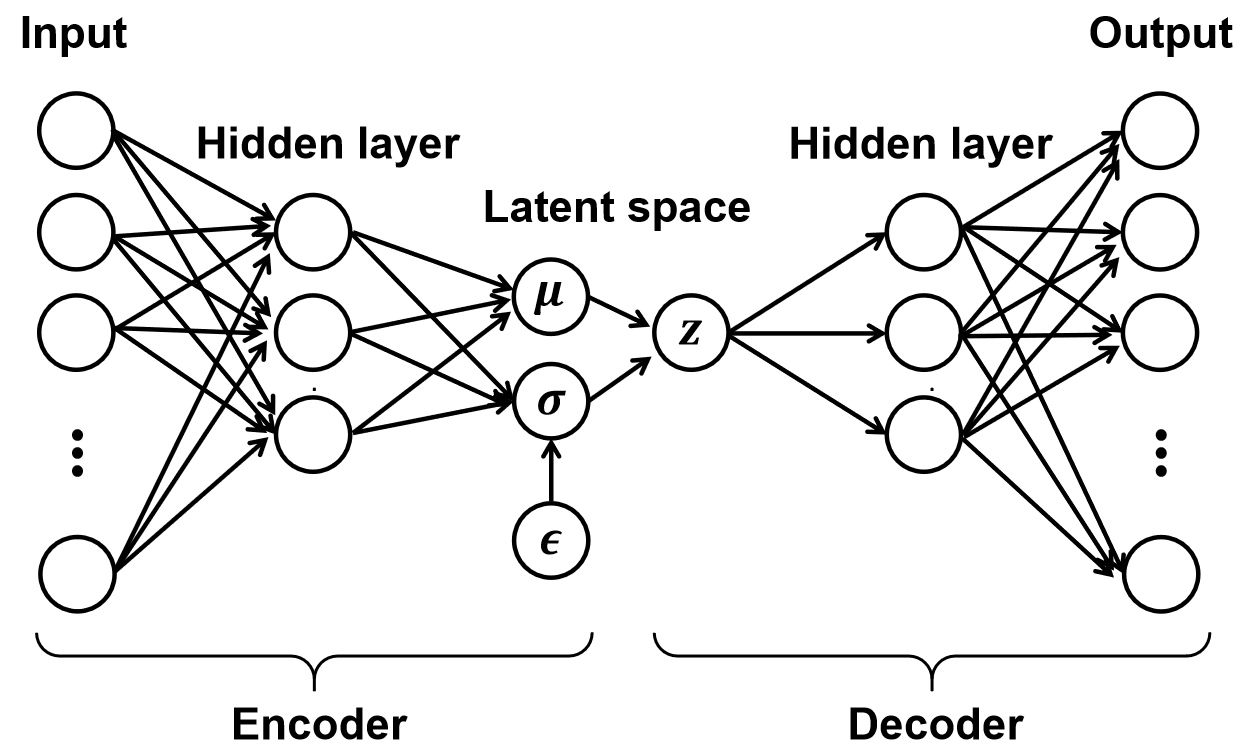}
\caption{Network architecture of the VAE}
\label{vae}
\end {center}
\end{figure}

\subsection{Variational autoencoder}\label{VAE}
In order to demonstrate the generality of the proposed approach, we constructed a simple VAE architecture to obtain all the results in this paper, avoiding shifting the focus to how to build the optimal network architecture.
The VAE consists of two main parts, encoder and decoder as shown in Fig.~\ref{vae}.
The encoder consists of input and hidden layers, and the number of neurons depends on the amount of input data.
\begin{figure*}[t]
\begin {center}
\includegraphics[width=1 \textwidth]{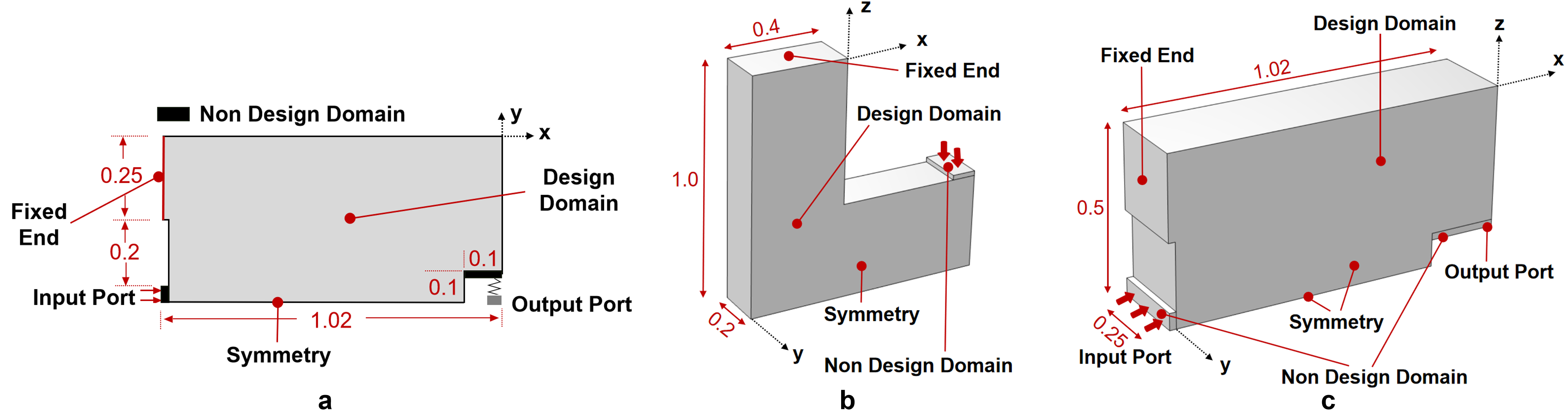}
\caption{Boundary conditions and design domains: \textbf{a} 2D compliant mechanism design problem, \textbf{b} maximum von Mises stress (MVMS) minimization problem, and \textbf{c} 3D compliant mechanism design problem}
\label{exam_domain}
\end {center}
\end{figure*}
It should be noted that a higher number of neurons in the hidden layers assumes a greater fitting and generating power which also results in a larger computational cost. 
Therefore, users usually need to make a trade-off between effectiveness and efficiency, and in the case of this paper two hidden layers are set up with the number of neurons 10000, 500 respectively.
After activating these neurons in hidden layers using the ReLU function,
this layer is also fully connected to two layers having 8 neurons, one corresponds to $\mu$, which is the mean value vector of the latent variables $\mathbf{z}$, and the other corresponds to $\log (\boldsymbol{\sigma} \circ \boldsymbol{\sigma})$), where $\boldsymbol{\sigma}$ is the variance vector of $\mathbf{z}$, and
$\circ$ represents the element-wise product.
We then obtain the latent variables $\mathbf{z} \in\mathbb{R}^{N_{ltn}}$ ($N_{\mathrm{ltn}}$ is the number of the latent variables) as follows
\begin{equation}
\label{eq7}
\mathbf{z}=\boldsymbol{\mu}+\boldsymbol{\sigma}  \boldsymbol {\epsilon} ,
\end{equation}
where $\boldsymbol{\epsilon}$ is a random vector according to the standard normal distribution.
The layer of the latent variables $\mathbf{z}$ is further fully connected to two hidden layers with the number of neurons 500, 10000 respectively.
The last hidden layer is further fully connected to the output layer without any activation such as the sigmoid activation.
This is because that the range of the principal component score is not limited to $[0, 1]$.
The VAE can generate meaningful new data using the decoder by imposing a regularization such that the compressed data are continuously distributed on a Gaussian in the latent space.
The VAE with the above architecture is trained using the elite data as the input data, and the latent space composed of the latent variables is constructed through training. 
In more detail, the training is conducted by minimizing the following loss function $L$ using the Adam optimizer \citep{kingma2014adam}:
\begin{equation}
\label{eq8}
 L:=L_{\mathrm{rcn}}+\beta L_{\mathrm{KL}} ,
\end{equation}
where $L_{\mathrm{rcn}}$ is the reconstruction loss measured by the mean-squared error, and $L_{\mathrm{KL}}$ is a term corresponding to
the Kullback-Leibler (KL) divergence. 
$\beta$ is the weighting coefficient for the KL divergence loss. 
$L_{\mathrm{KL}}$ is computed as follows:
\begin{equation}
\label{eq9}
L_{\mathrm{KL}}=-\frac{1}{2} \sum_{i=1}^{N_{\mathrm{ltn}}}\left(1+\log \left(\sigma_{i}^{2}\right)-\mu_{i}^{2}-\sigma_{i}^{2}\right) ,
\end{equation}
where $\mu_{i}$ and $\sigma_{i}$ are the $i$-th components of $\mu$ and $\sigma$, respectively.

\section{Experiment}\label{sec4}
In this section, we conduct several numerical experiments to validate the effectiveness of the proposed PCA-based DDTD.
All experiments are conducted on a computer with Linux x86$\_$64 architecture and 128 cores.

\subsection{Problem settings} \label{pro}
TO methods which target geometrically linear problems under the assumption of small deformations are not applicable to more complex practical applications under large deformations, such as energy absorbing structures, compliant mechanisms.
Thus, geometric nonlinearity is considered in this paper to obtain a more realistic design for applications with large deformations.

More specifically, in Sec~\ref{exam_comp} and Sec~\ref{exam2}, we select the followings as the objective functions in Eq.~(\ref{eq1}) to obtain geometric nonlinear compliant mechanism structures:
\begin{equation}
\label{eq2}
\left\{
\begin{array}{l}
J_{1}=\max \left(\sigma_{\text{v}}\right) \\ \\
J_{2}=V \\ \\
J_{3}=-F_{\text{r}}
\end{array}\right.,
\end{equation}
where, $\sigma_{\text{v}}$ is the von Mises stress in the structure, $V$ is the volume of the structure, and $F_{\text{r}}$ is the reaction force yielded by an artificial spring set on the output port. 
The negative sign of $F_{\text{r}}$ is needed to convert the maximization problem to a minimization problem.
Hereafter, we represent the reaction force with $J_3$, that is, reaction force values are displayed in negative values.
We don't consider constraint functions here, therefore, $N_{\text{cns}}$ is 0.

In Sec~\ref{exam1}, we select the following objective functions to obtain structures whose maximum von Mises stress (MVMS) is low:
\begin{equation}
\label{eq2_add}
\left\{
\begin{array}{l}
J_{1}=\max \left(\sigma_{\text{v}}\right) \\ \\
J_{2}=V
\end{array}\right.,
\end{equation}
where $N_{\text{cns}}$ is 0.

It should be noted that due to the localization, singularity, non-linearity, and metric accuracy of the stress, traditional TO methods circumvent the direct solving via approximating the original problem.
In this research, all objective functions including the MVMS are accurately calculated using a body-fitting mesh while avoiding problems such as accuracy loss caused by the approximation approaches.

\begin{figure}[t]
\begin {center}
\includegraphics[width=0.48 \textwidth]{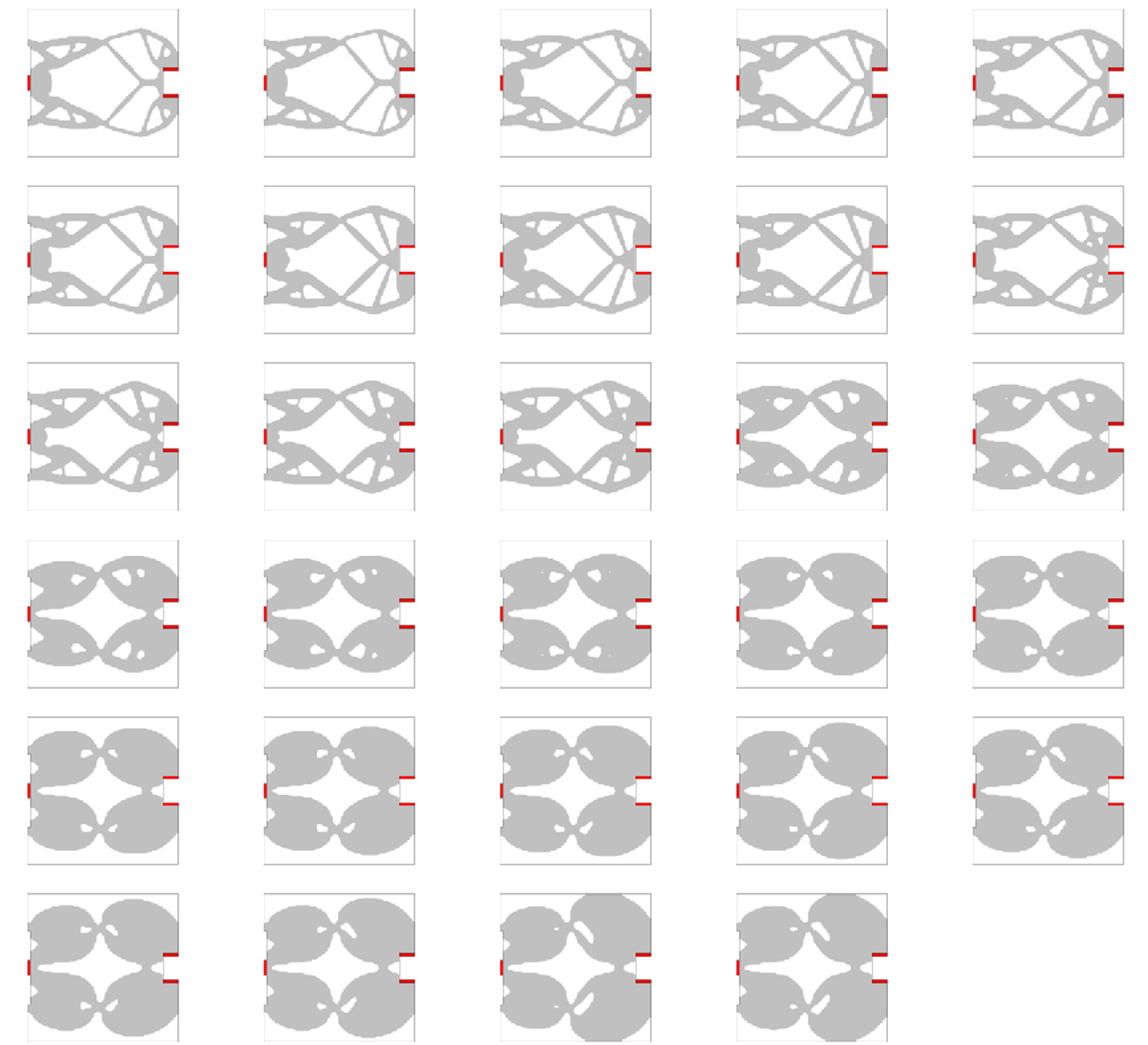}
\caption{Elite material distributions at iteration~0 in comparison with Non PCA-based DDTD}
\label{comp_initial}
\end {center}
\end{figure}

\begin{figure}[t]
\begin {center}
\includegraphics[width=0.45 \textwidth]{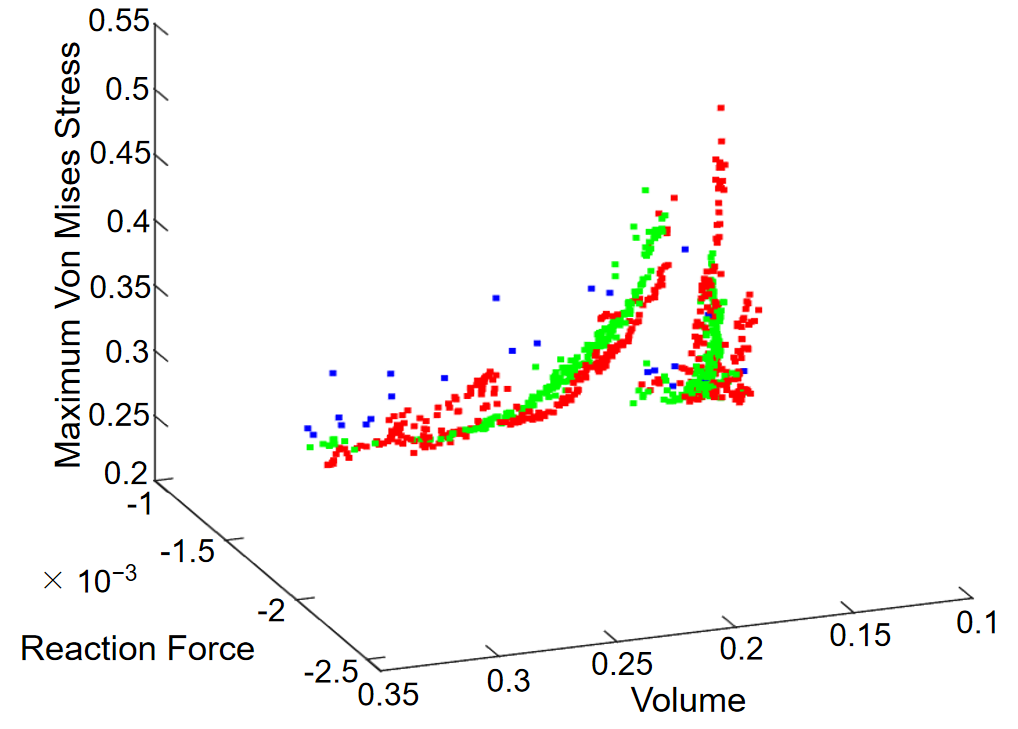}
\caption{Performances of elite material distributions in comparison with Non PCA-based DDTD: iteration~0 (blue), iteration~50 in PCA-based DDTD (red), and iteration~50 in Non PCA-based DDTD (green)}
\label{Comparison_pareto}
\end {center}
\end{figure}

\begin{figure}[t]
\begin {center}
\includegraphics[width=0.45 \textwidth]{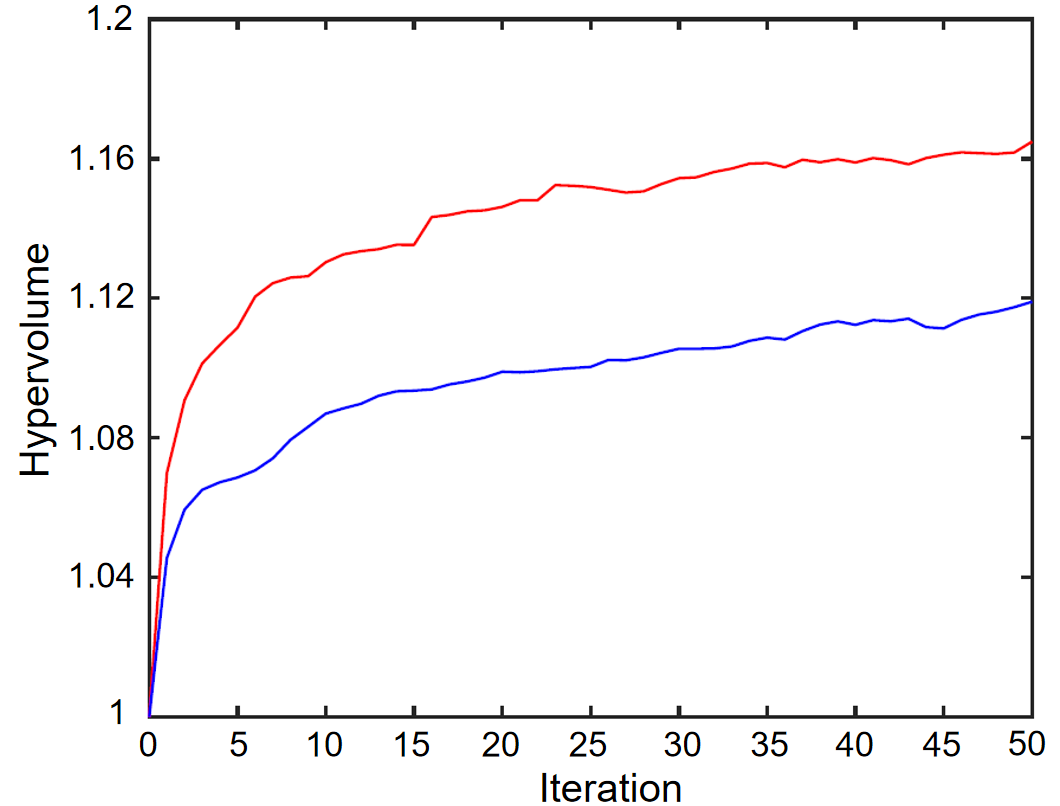}
\caption{History of hypervolume: Non PCA-based DDTD (blue) and PCA-based DDTD (red)}
\label{Comparison_hyper}
\end {center}
\end{figure}

\begin{figure*}[t]
\begin {center}
\includegraphics[width=1 \textwidth]{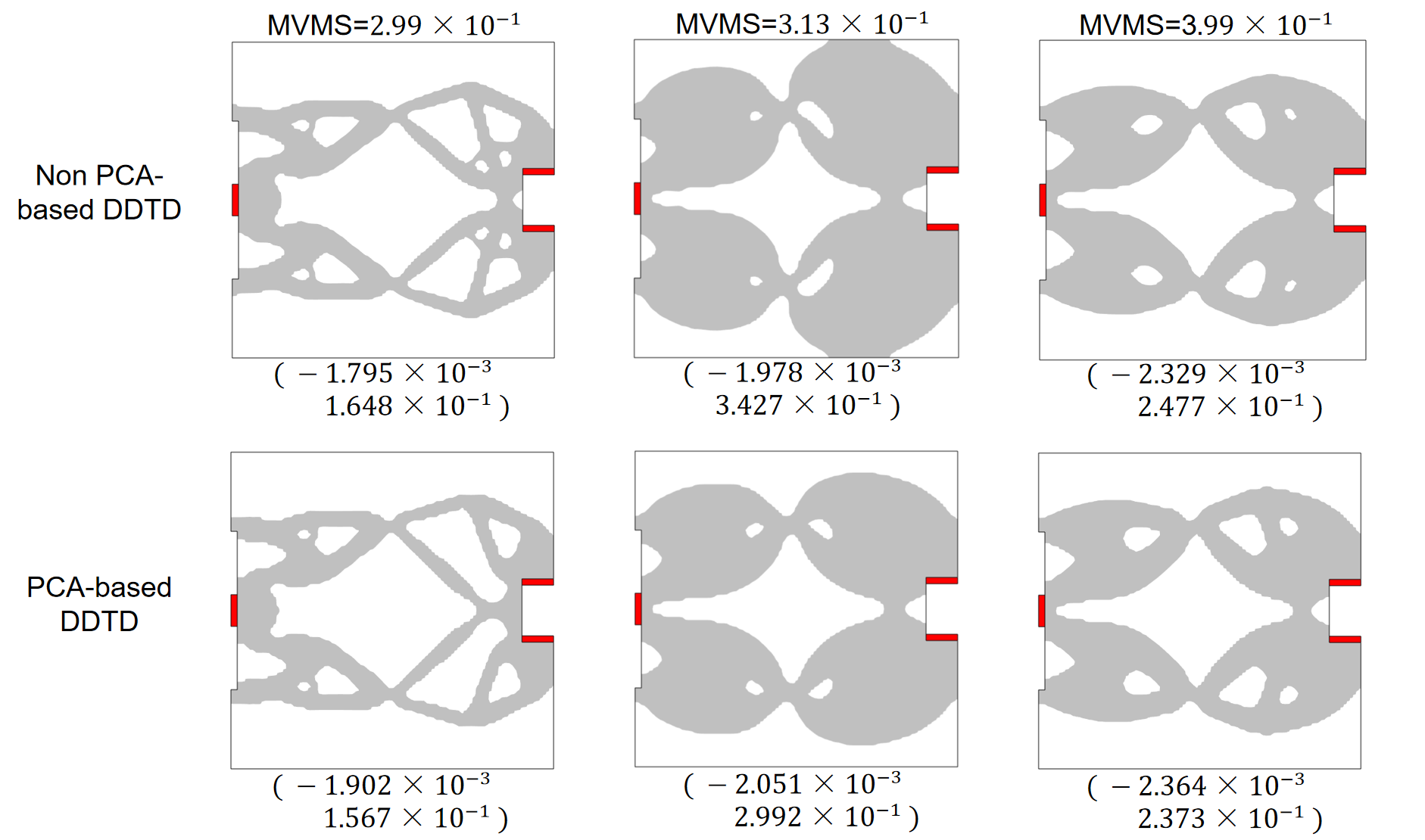}
\caption{Comparison of elite material distributions at iteration~50 in PCA-based and Non PCA-based DDTDs. Here, the left and right sides of the bracket indicate the reaction force and volume values, respectively}
\label{Comparison_results}
\end {center}
\end{figure*}

\begin{figure*}[t]
\begin {center}
\includegraphics[width=1 \textwidth]{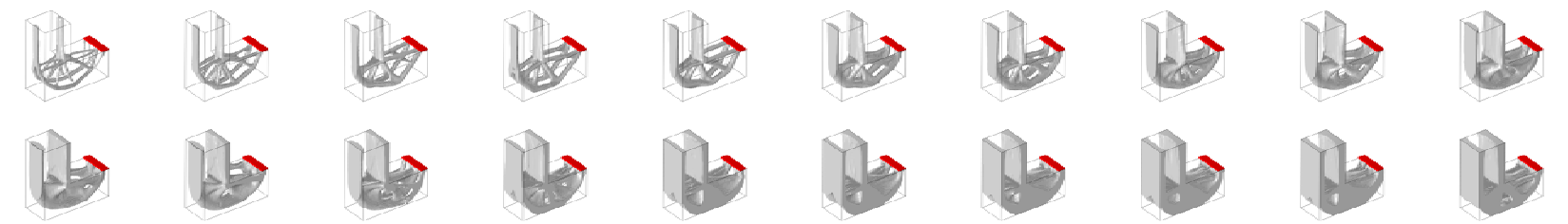}
\caption{Part of elite material distributions at iteration~0 in MVMS minimization problem without considering large deformation }
\label{L_nonlarge_initial 41}
\end {center}
\end{figure*}

\begin{figure*}[t]
\begin {center}
\includegraphics[width=0.9 \textwidth]{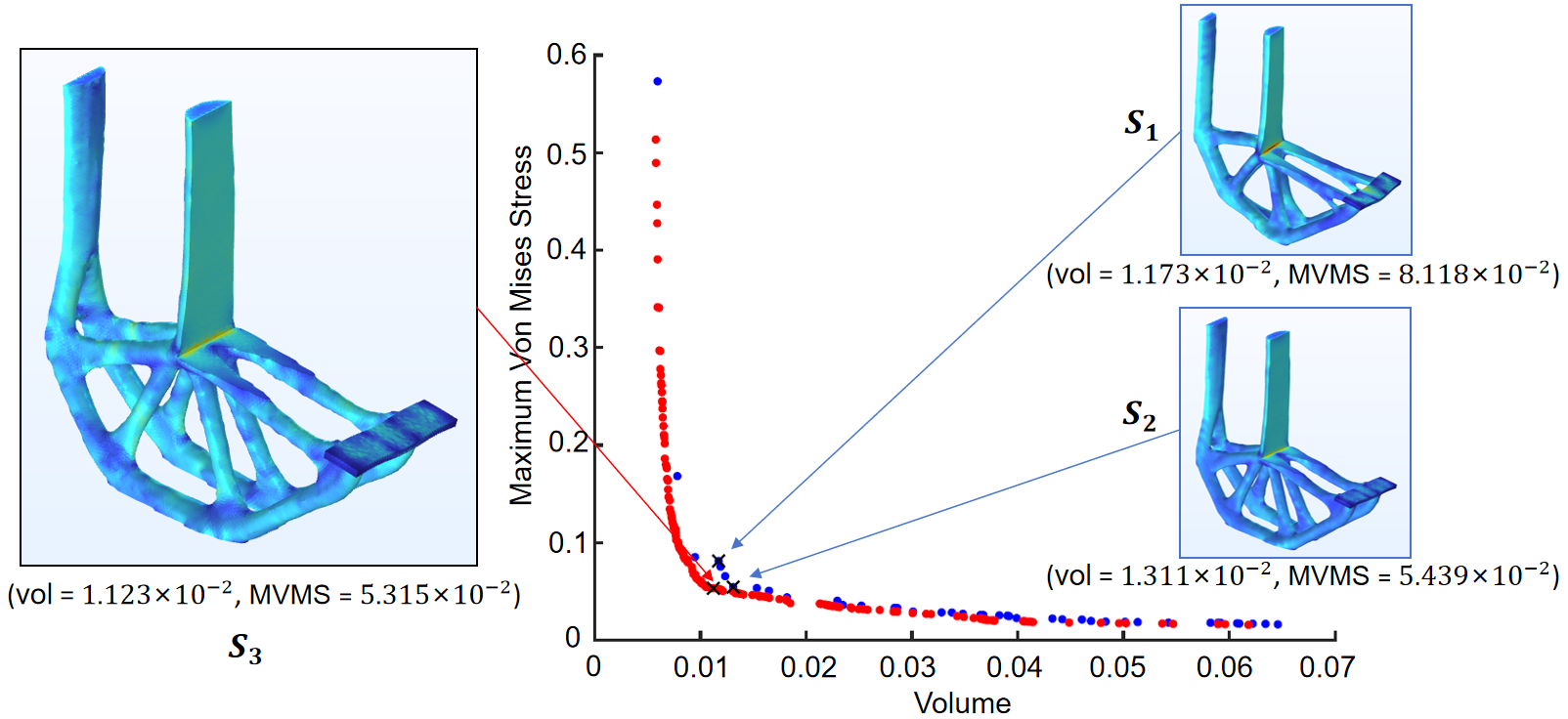}
\caption{Performances of elite material distributions in MVMS minimization problem without considering large deformation: iteration~0 (blue) and iteration~50 (red). Here, \textsf{vol} means volume}
\label{L_nonlarge par}
\end {center}
\end{figure*}

\begin{figure*}[t]
\begin {center}
\includegraphics[width=1 \textwidth]{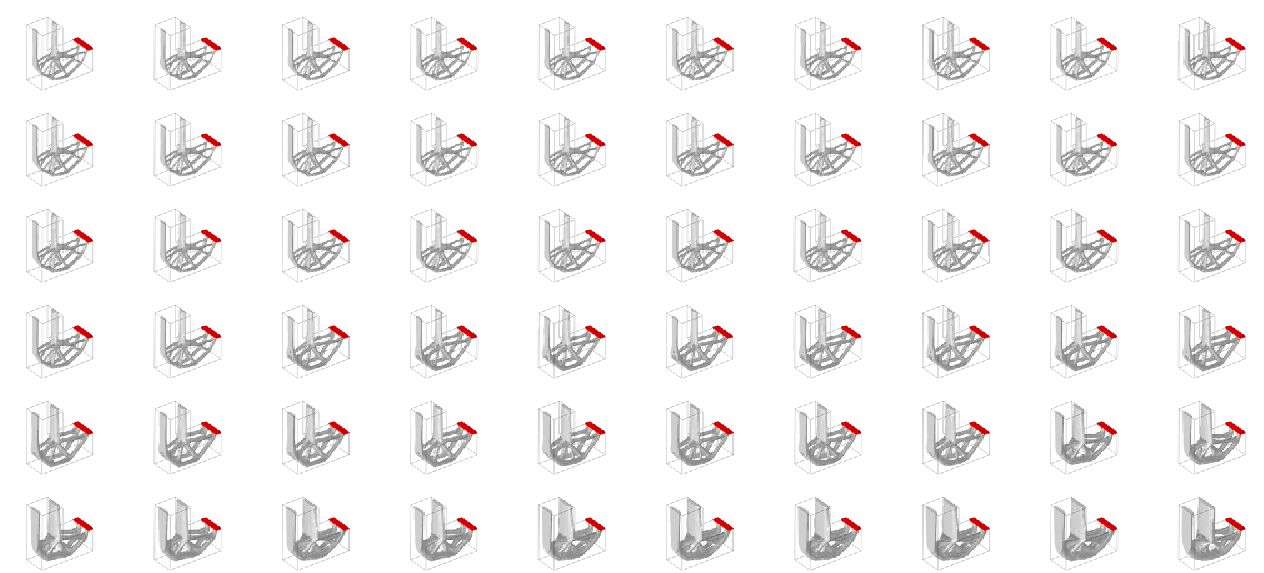}
\caption{Part of elite material distributions at iteration~50 in MVMS minimization problem without considering large deformation }
\label{L_nonlarge_results149}
\end {center}
\end{figure*}

\begin{figure*}[t]
\begin {center}
\includegraphics[width=1 \textwidth]{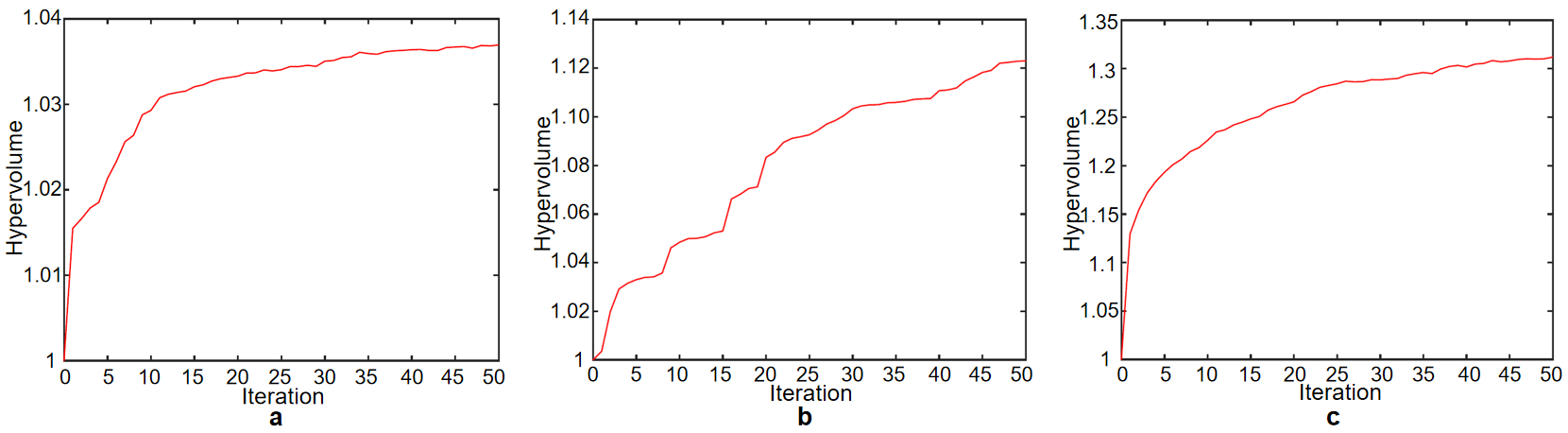}
\caption{History of hypervolumes: \textbf{a} MVMS minimization problem without considering large deformation, \textbf{b} MVMS minimization problem with considering large deformation, and \textbf{c} 3D compliant mechanism design problem}
\label{Hypervolume}
\end {center}
\end{figure*}

\begin{figure}[t]
\begin {center}
\includegraphics[width=0.5 \textwidth]{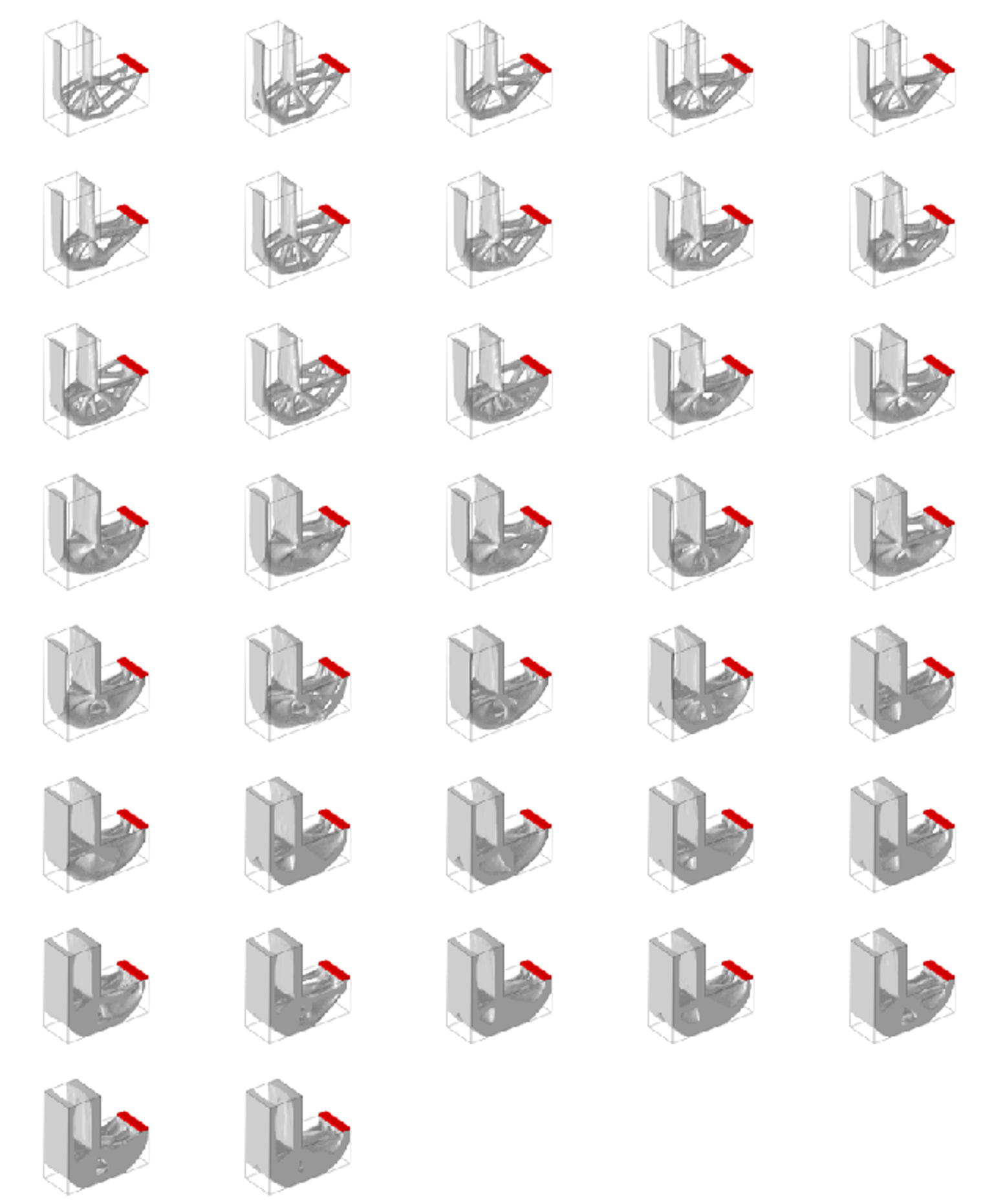}
\caption{Elite material distributions at iteration~0 in MVMS minimization problem with considering large deformation}
\label{L_large_initial 37}
\end {center}
\end{figure}

\begin{figure*}[t]
\begin {center}
\includegraphics[width=0.8 \textwidth]{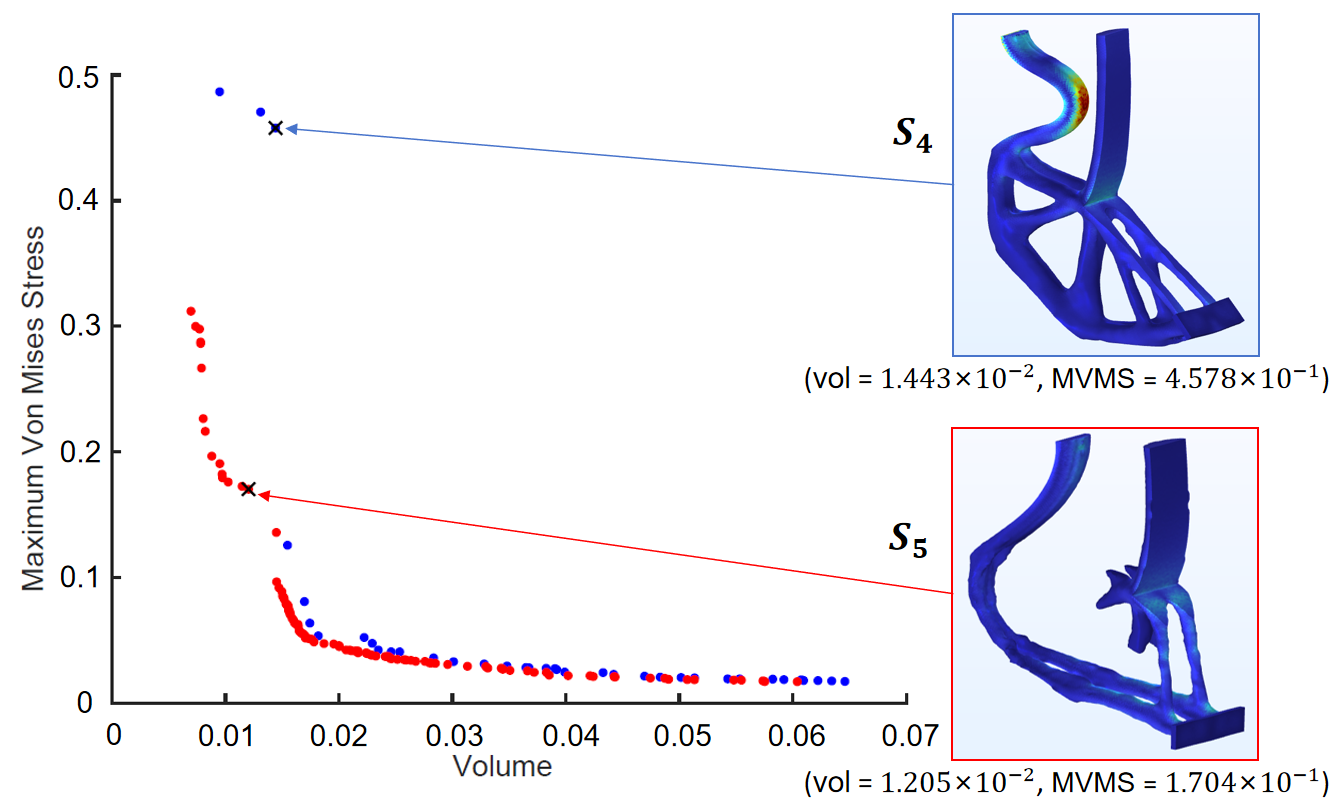}
\caption{Performances of elite material distributions in MVMS minimization problem with considering large deformation: iteration~0 (blue) and iteration~50 (red). Here, \textsf{vol} means volume}
\label{L_large par}
\end {center}
\end{figure*}

\begin{figure*}[t]
\begin {center}
\includegraphics[width=1 \textwidth]{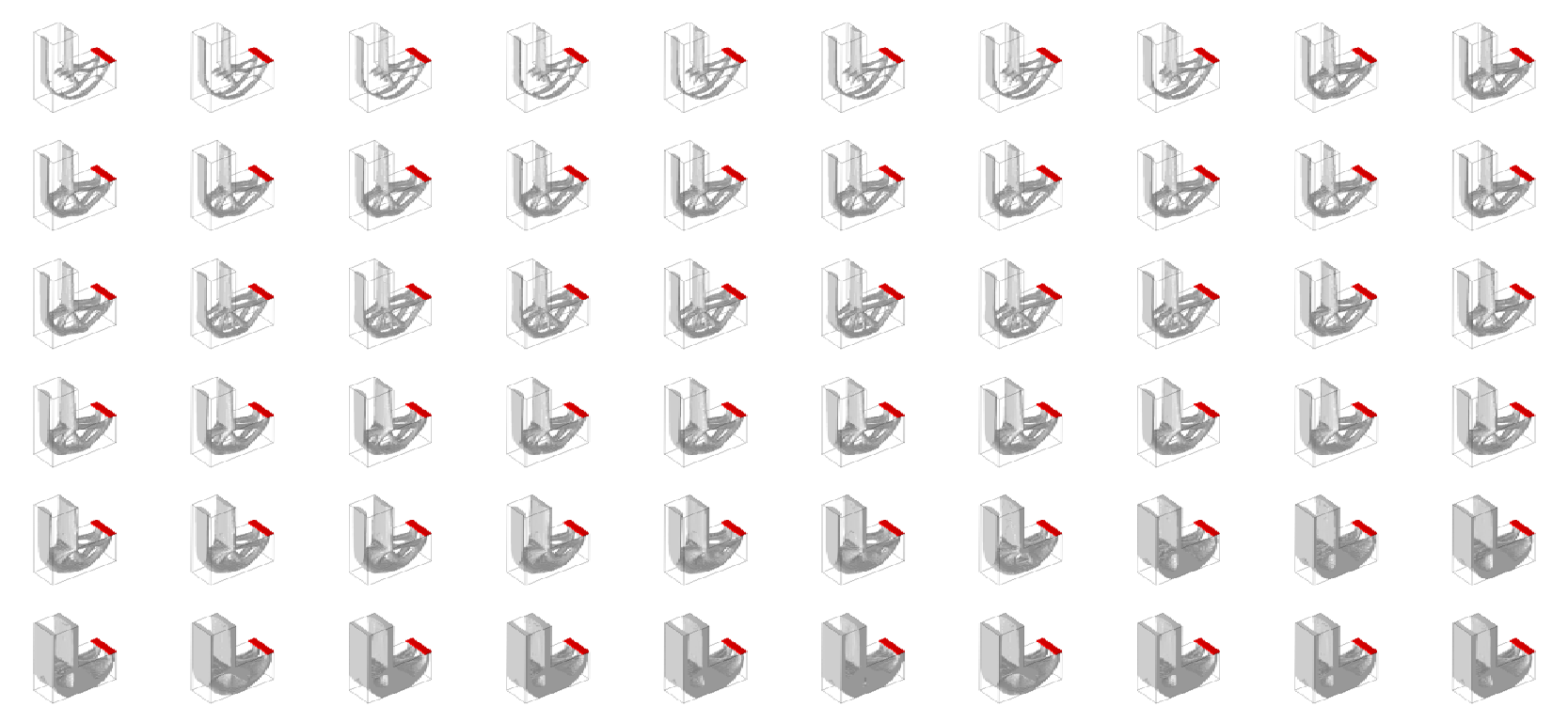}
\caption{Part of elite material distributions at iteration~50 in MVMS minimization problem with considering large deformation}
\label{L_nonlarge_results111}
\end {center}
\end{figure*}

\begin{figure}[t]
\begin {center}
\includegraphics[width=0.45 \textwidth]{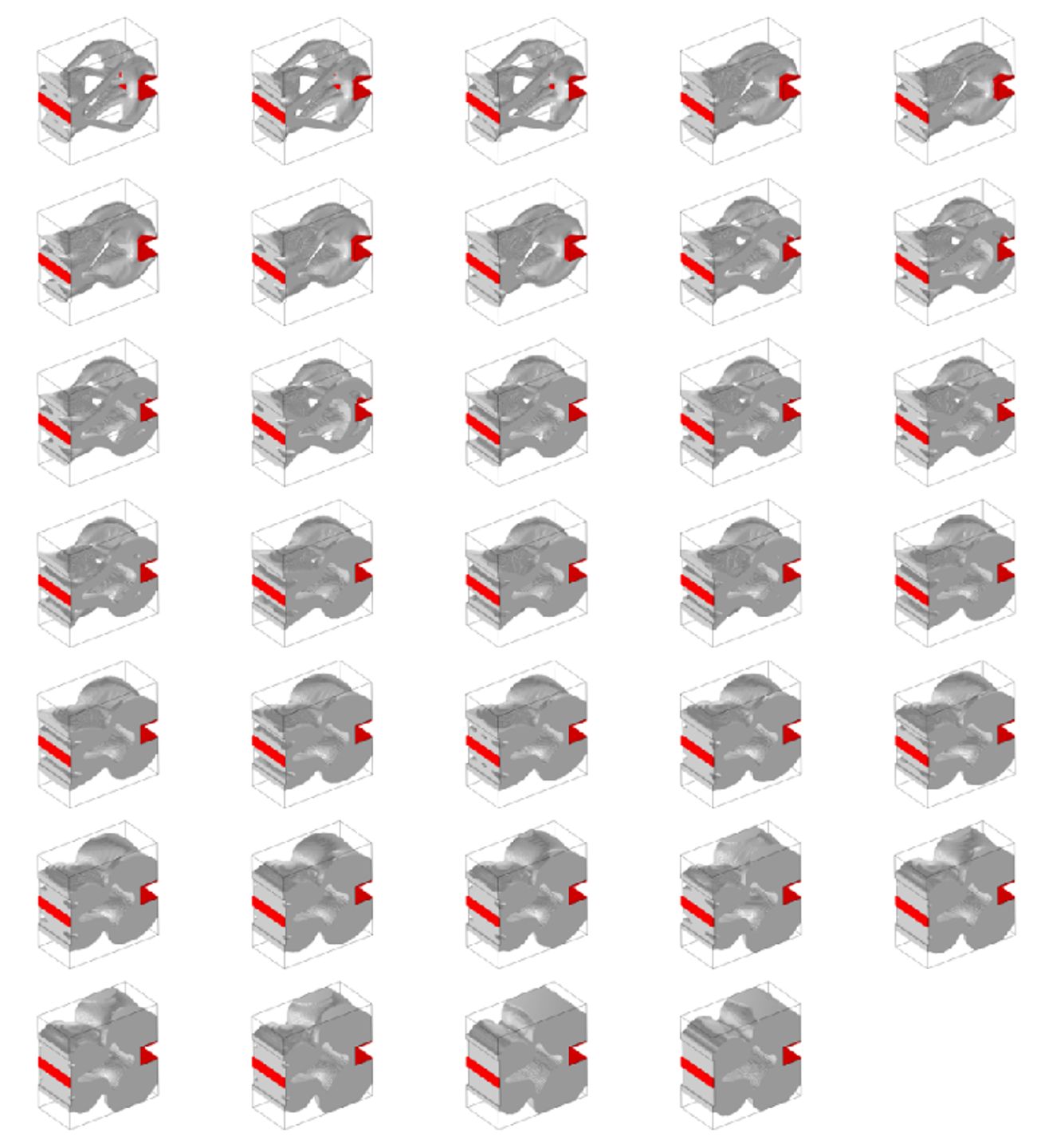}
\caption{Part of elite material distributions at iteration~0 in 3D compliant mechanism design problem}
\label{Compliant_initial}
\end {center}
\end{figure}

\begin{figure*}[t]
\begin {center}
\includegraphics[width=1 \textwidth]{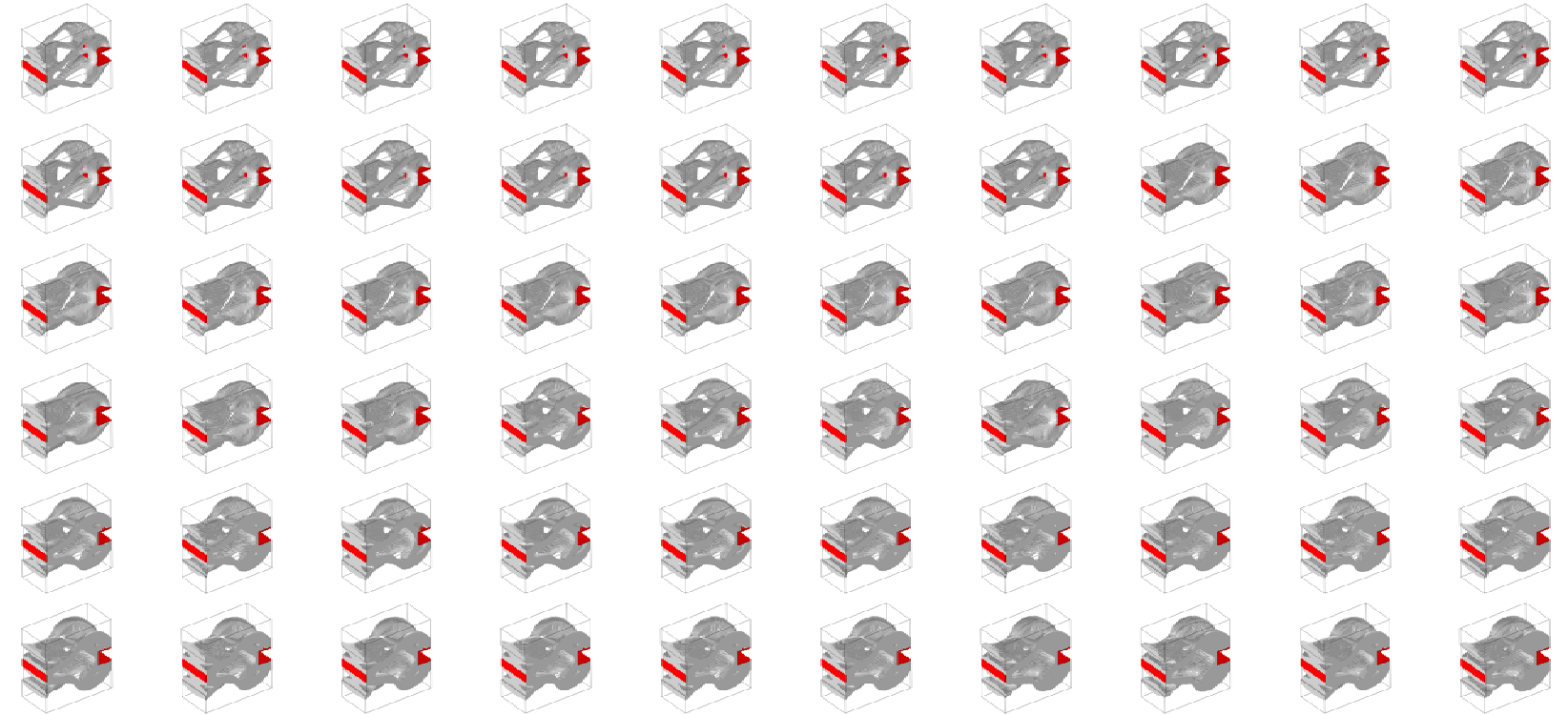}
\caption{Part of elite material distributions at iteration~50 in 3D compliant mechanism design problem}
\label{Compliant_results400}
\end {center}
\end{figure*}

\begin{figure}[t]
\begin {center}
\includegraphics[width=0.48 \textwidth]{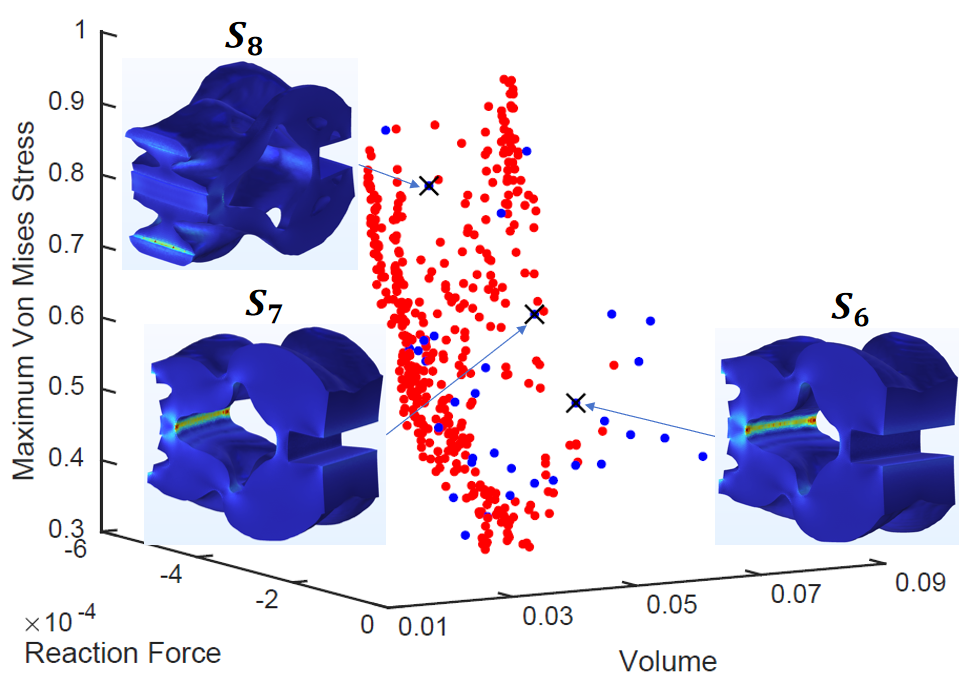}
\caption{Performances of elite material distributions in 3D compliant mechanism design problem: iteration~0 (blue) and iteration~50 (red)}
\label{Compliant_par}
\end {center}
\end{figure}

\begin{figure*}[t]
\begin {center}
\includegraphics[width=1 \textwidth]{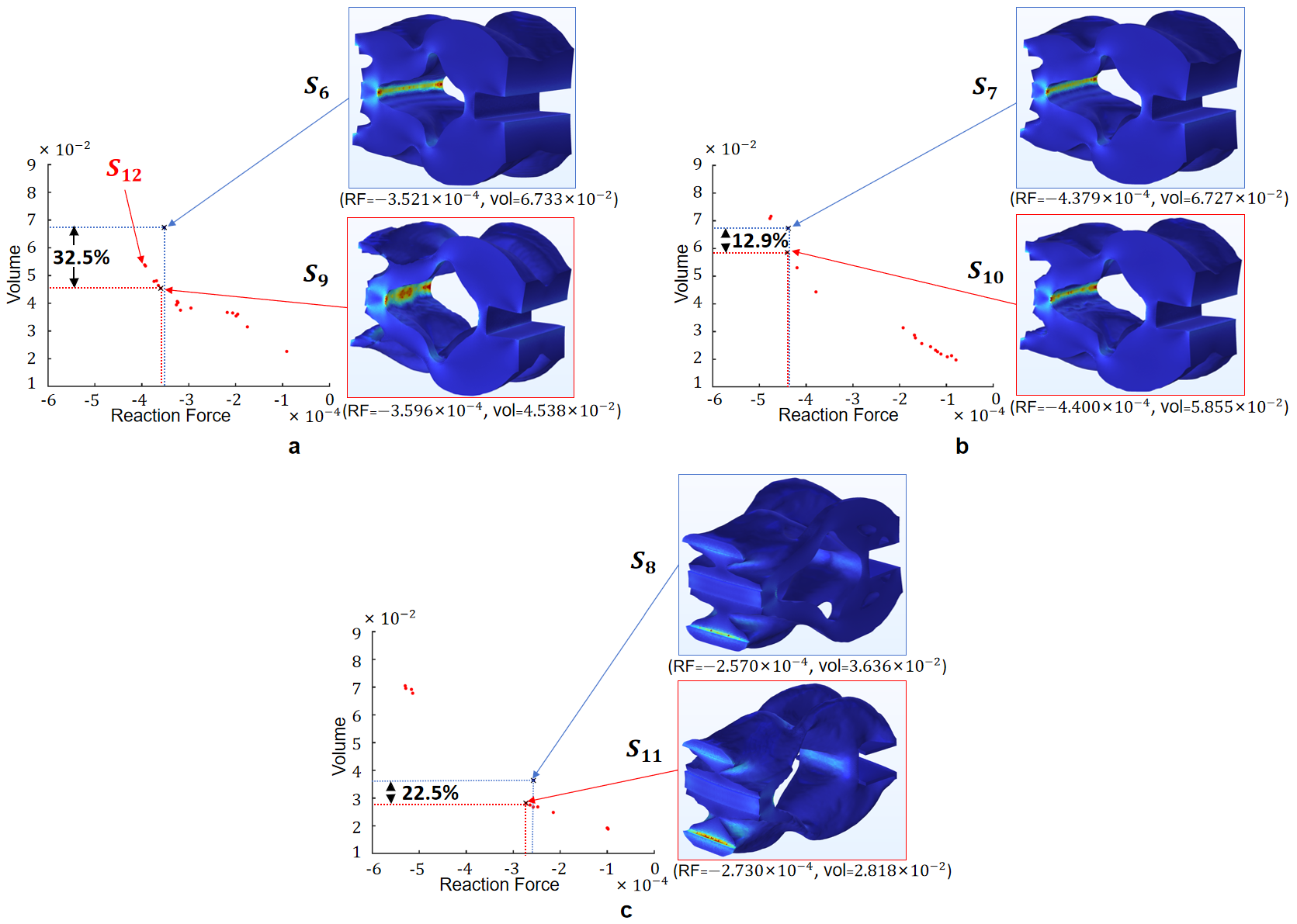}
\caption{Comparison of elite material distributions at iterations~0 (blue) and 50 (red) in 3D compliant mechanism design problem: \textbf{a} comparison under MVMS $\in 0.4800 \pm 0.01$, \textbf{b} comparison under MVMS $\in 0.5891 \pm 0.01$, and \textbf{c} comparison under MVMS $\in 0.8252 \pm 0.01$. Here, \textsf{RF} and \textsf{vol} mean reaction force and volume, respectively}
\label{Compliant_comp}
\end {center}
\end{figure*}

As common settings, we use the following for the training of the VAE:
\begin{itemize}
    \item {learning rate}: $1.0\times  10^{-4} $
    \item {mini-batch size}: $20$
    \item {number of epochs}: $400$
    \item {number of latent variables, $N_{\text{ltn}}$}: $8$
    \item {weighting coefficient for KL divergence, $\beta$}: $4$
\end{itemize}
In addition, we use the latent crossover proposed by~\citet{Kii2024latent} for the sampling in the latent space.
Other parameters in the DDTD process are set as follows:
\begin{itemize}
    \item {maximum number of elite data}: $400$
    \item {total number of iterations}: $50$
\end{itemize}

\subsection{Comparison} \label{exam_comp}
To verify the effectiveness of the proposed PCA-based DDTD, we compare our results with DDTD without PCA (Non PCA-based DDTD), in which the nodal densities are directly used as the training data for the VAE.
Due to the limitation on input data size for deep generative models, it is challenging to apply Non PCA-based DDTD to optimization problems with massive DOFs, particularly in 3D structural design.
Therefore, we compare the proposed PCA-based DDTD with Non PCA-based DDTD in the setting of a 2D compliant mechanism design problem shown in Fig.~\ref{exam_domain}(a).
As shown here, the numerical analysis is done using the half model, a horizontal load of $0.08$ is applied on the input port, an artificial spring of $10$ is set on the output port, and the displacement is fixed on the fixed end.
Young's modulus and Poisson's ratio are set to $1$ and $0.3$, respectively.

To ensure fairness, we compare PCA-based and Non PCA-based DDTDs under the same conditions except for two points.
First, we changed the neural network architecture of the VAE used in Non PCA-based DDTD because the size of the input and output layers are quite different. 
Those size is equal to the number of the nodal densities $n$ in Non PCA-based DDTD, and the number of the intermediate layer is one, whose size is 500, for both the encoder and decoder.
Further, we use the sigmoid activation before the output layer because the outputs are the nodal densities in this case.
Next, in Non PCA-based DDTD, material distributions are normalized with $h = 0.02$ for training the VAE.
This is because that it is suggested to normalize the material distributions with large $h$ in one representative implementation of Non PCA-based DDTD~\citep{yamasaki2021data}.

Initial material distributions are obtained by solving a low-fidelity problem using a density-based TO method, which is easily and directly solvable, yet relevant to the original multi-objective optimization problem.
More specifically, the low-fidelity problem does not consider the large deformation and replaces minimizing the MVMS with limiting the amount of the deformation on the input port to reduce computational complexity.
For these initial material distributions, we evaluate the objective function values of the original problem.
By doing so, we obtain 29 initial elite material distributions shown in Fig.~\ref{comp_initial}.

Figure~\ref{Comparison_pareto} shows how the performances of the elite material distributions are improved in this example.
As shown in this figure, the variation of the elite solutions towards the minimization in the objective function space is more obvious in PCA-based DDTD compared to Non PCA-based DDTD, which proves that PCA-based DDTD has better effectiveness.

In order to more directly compare the difference in effectiveness between PCA-based and Non PCA-based DDTDs, we evaluate the performance of the whole elite solutions for each iteration using the hypervolume indicator, which measures the whole performance of the multi-objective optimization \citep{Kii2024latent}.
We evaluate and compare them using the hypervolume indicator, which is normalized using the initial value. 
When the value of hypervolume indicator is greater than 1, it indicates that the elite solutions are progressing relative to the reference point.
In addition, a larger value of hypervolume indicator indicates a better performance compared to the reference point.
As shown in Fig.~\ref{Comparison_hyper}, the blue and red lines represent the hypervolume indicator of Non PCA-based DDTD and PCA-based DDTD, respectively.
We can observe that the hypervolume indicator improves throughout the entire iteration process, and both values are greater than 1, indicating that the performance of the results from both DDTDs is better than that of the initial elite material distributions.
It should be noted that the hypervolume indicator of PCA-based DDTD outperforms the hypervolume indicator of Non PCA-based DDTD throughout, which suggests that PCA-based DDTD is significantly more effective than Non PCA-based DDTD under a fair comparison.

Part of the finally obtained elite material distributions in PCA-based and Non PCA-based DDTDs are shown in Fig.~\ref{Comparison_results}.
Since this 2D compliant mechanism design problem contains three optimization objectives (volume, reaction force, MVMS), we chose three pairs of results under almost same MVMS (difference within $5 \times 10^{-4}$) for straightforward comparison.
As we have seen, the structural performance of the results of PCA-based DDTD is significantly better than the structural performance of the results of Non PCA-based DDTD (i.e., having smaller objective function values in both reaction force and volume).

\subsection{Numerical example 1} \label{exam1}

Here, we examine the popular L-shaped beam for testing under the condition of not considering and considering large deformations respectively.
In this optimization problem, as shown in Fig.~\ref{exam_domain}(b), the number of design variables is reduced to half of that in the original design domain, i.e., $138621$, due to the presence of symmetric boundary conditions about the $xz$ plane.
As mentioned earlier, material distributions at this scale cannot be learned efficiently by the VAE due to the presence of input size limitation.
With the benefit of PCA, the proposed PCA-based DDTD can resolve the conflict between the input size limitation of the VAE and the representation of complex structures.
For other problem settings, a vertical load of $0.002$ is applied on the tip of the L-shaped design domain, and Young's modulus and Poisson's ratio are set to $1$ and $0.3$, respectively.

In the optimization problem without considering large deformation, the stiffness of the structure is considered to be independent of the magnitude of the applied loading and the displacement matrix is unique.
We firstly obtained initial material distributions by solving a low-fidelity optimization problem using a density-based TO method, which utilizes the approximation based on the $P$-norm and essentially the same as the problem solved by~\citet{Kii2024latent}.
After that, we selected $41$ initial elite material distributions shown in Fig.~\ref{L_nonlarge_initial 41}.

Figure~\ref{L_nonlarge par} shows how the performances of the elite material distributions are improved through the DDTD iterative process.
As shown in this figure, the elite solutions are migrating towards the minimization in the objective function space, which proves that the performances of the elite material distributions are improving via DDTD process.
We here chose two structures at iteration~0, $S_1$ and $S_2$ (blue boxes), and one structure at iteration~50, $S_{3}$ (red box), in the volume range of $[0.01, 0.015]$, to explain the changes in the shape and topology of material distributions during the generation process of DDTD.

Compared to $S_1$, which has a lower volume, $S_2$ has more branches to spread out the stresses, thus leading to the advantages of $S_1$ and $S_2$ in terms of volume and stress, respectively.
$S_{3}$ inherits the multiple branches in $S_2$ and its shape is modified to widely distribute the von Mises stress through the DDTD process, thus outperforming $S_1$ and $S_2$ in terms of the volume ($S_1$, $S_2$, $S_{3}$ are 1.173, 1.311, 1.123 $\times 10^{-2}$ respectively) and the stress ($S_1$, $S_2$, $S_{3}$ are 8.118, 5.439, 5.315 $\times 10^{-2}$ respectively).
Figure~\ref{L_nonlarge_results149} shows a part of the elite material distributions at iteration~50.
As shown in Fig.~\ref{Hypervolume}(a), the hypervolume indicator is progressing throughout the iterations, which indicates that the whole performance of the solutions is improving.

In the optimization problem with considering large deformation, the deformation of the structure by the applied load can not be neglected, which means that the stiffness matrix of the structure is changed along with the amount of the deformation and nodal displacements. 
It should be noted that considering large deformations is more meaningful in engineering problems but it also causes more computational complexity.

From the same initial material distributions to the case of without considering large deformation, we select $37$ initial elite material distributions shown in Fig.~\ref{L_large_initial 37}.
Figure~\ref{L_large par} shows how the performances of the elite material distributions are improved through the DDTD iterative process.
In this figure, there is a significant stress concentration in the structure at iteration~0, $S_{4}$, which can easily lead to extreme fragility of the structure.
While it seems that the multiple branches appearing in $S_{4}$ are effective in spreading the stresses, the branches with high stiffness existing in the lower half of the design domain are conversely the most significant factor contributing to the problem of the localized high stress concentration.
In the structure at iteration~50, $S_{5}$, this localized stress concentration problem is avoided by removing those seemingly effective high stiffness structures.

Part of the elite material distributions at iteration~50 are shown in Fig.~\ref{L_nonlarge_results111}.
It should be noted that these new features appearing in the elite material distributions at iteration~50 do not exist in those at iteration~0, which proves that the DDTD process not only inherits the features of the initial data, but also brings in brand novel features to play the role of optimization.
As a result, $S_{5}$ is significantly better than $S_{4}$ in terms of both volume ($S_{4}$, $S_{5}$ are 1.443, 1.205 $\times 10^{-2}$ respectively) and stress ($S_{4}$, $S_{5}$ are 4.578, 1.704 $\times 10^{-2}$ respectively), and even has an 62.9\% decrease in stress.
As shown in Fig.~\ref{Hypervolume}(b), the hypervolume indicator is approximately 1.122 at iteration~50, which indicates that the whole performance of the elite material distributions is improving.

\subsection{Numerical example 2}\label{exam2}

In order to further verify the validity of the proposed PCA-based DDTD, we chose 3D compliant mechanism design problem shown in Fig.~\ref{exam_domain}(c).
In this optimization problem, the number of design variables is reduced to one-fourth of that in the original design domain, i.e., $130078$, due to the presence of the symmetric boundary conditions about the $xy$, $xz$ plane.
As previously mentioned, Non PCA-based DDTD cannot directly use material distribution data of this scale to train the VAE due to the limitation of the maximum input size.
For other problem settings, a horizontal load of $0.08$ is applied on the input port, an artificial spring of $10$ is set on the output port, and the displacement is fixed on the fixed end.
Young's modulus and Poisson's ratio are set to $1$ and $0.3$, respectively.

Similar to the case of the 2D compliant mechanism design problem, initial material distributions are obtained by solving a low-fidelity problem using a density-based TO method.
After that, we evaluate the objective function values of the original problem.
By doing so, we obtain $29$ initial elite material distributions shown in Fig.~\ref{Compliant_initial}.

Starting from these elite material distributions, we finally obtain the elite material distributions shown in Fig.~\ref{Compliant_results400}.
Figure~\ref{Compliant_par} also shows how the performances of the elite material distributions are improved through the DDTD iterative process.
In this figure, the elite material distributions are migrating towards the minimization in the objective function space, which proves that the performances of the elite material distributions are improving during the iterations.
As shown in Fig.~\ref{Hypervolume}(c), the hypervolume indicator is approximately 1.31 at iteration 50, which indicates that the whole performance of the elite material distributions has improved significantly.

In order to provide a more obvious comparison of the performance difference between the elite material distributions at iterations~0 and 50, we choose three structures at iteration~0 as a benchmark, denoted as $S_6$, $S_7$, and $S_8$, respectively.
As shown in Fig.~\ref{Compliant_comp}(a), PCA-based DDTD generates diverse material distributions with higher performances than the benchmark $S_6$ at similar MVMS (MVMS $\in 0.4800\pm0.01$, $S_6$ and $S_9$ are 0.4800, 0.4731, respectively).
The newly generated structure $S_9$ slightly outperforms $S_6$ in terms of the reaction force ($S_6$ and $S_9$ are $-3.521$ $\times 10^{-4}$, $-3.596$ $\times 10^{-4}$, respectively) while significantly outperforming $S_6$ in terms of the volume ($S_6$ and $S_9$ are $6.733$ $\times 10^{-2}$, $4.538$ $\times 10^{-2}$, respectively), with a decrease of $32.5\%$, proving that lightweighting is effectively achieved while other properties remain similar.
Compared to $S_6$, whose stress is concentrated at the ends of the bar at the input side, the stress in $S_9$ is relatively uniformly distributed over the entire bar, which demonstrates that the stress concentration of the structure is effectively solved in PCA-based DDTD, and this is the main reason why the volume of the material decreases significantly while the structure's performance remains unchanged.

In Fig.~\ref{Compliant_comp}(b), we selected material distributions whose MVMS $\in 0.5891\pm0.01$, for comparison.
As shown in this figure, PCA-based DDTD generates new material distributions with diverse structural properties compared to the benchmark $S_7$.
We choose $S_{10}$, which has similar properties to $S_7$ in terms of the reaction force ($S_7$ and $S_{10}$ are $-4.379$ $\times 10^{-4}$ , $-4.400$ $\times 10^{-4}$, respectively), for illustration.
As a result, we obtain the same conclusion as in Fig.~\ref{Compliant_comp}(a), i.e., by eliminating the stress concentration of the initial structure, PCA-based DDTD can achieve the effect of maintaining the other structural properties unchanged while effectively reducing the material usage ($S_7$ and $S_{10}$ are $6.727$ $\times 10^{-2}$, $5.855$ $\times 10^{-2}$, respectively, decreasing $12.9\%$).

In Fig.~\ref{Compliant_comp}(c), we selected material distributions whose MVMS $\in 0.8252\pm0.01$, for comparison.
In order to better show the stress concentration part, we changed the viewpoint from that of Fig.~\ref{Compliant_comp}(a)-(b).
We selected structure $S_{11}$ because it has a similar value in terms of the reaction force ($S_8$ and $S_{11}$ are $-2.570 \times 10^{-4}$, $-2.730 \times 10^{-4}$, respectively).
The volumes of both are $3.636 \times 10^{-2}$ and $2.818 \times 10^{-2}$ (with a decrease of $22.5\%$), which again validate the aforementioned conclusions.

It should be noted that the focus on the volume comparison does not mean that DDTD only improves the initial material distributions in terms of material usage (lightweighting), rather DDTD can improve the initial material distributions on all the set optimization objectives to generate new material distributions with a diversity of structural properties.
For example, in Fig.~\ref{Compliant_comp}(a), newly generated structure $S_{12}$ outperformed benchmark $S_6$ obviously in both the reaction force and volume aspects, due to the smaller values of the optimization objectives.

\section{Conclusions} \label{sec5}

In this paper, we proposed PCA-based DDTD for solving the input limitation problem of the VAE.
In the proposed PCA-based DDTD, the VAE is trained by replacing the original material distributions with the principal component score matrix obtained by using PCA.
The material distributions with new features were generated by PCA-based DDTD.
This ensures that a high DOF of the material distribution representation is maintained while still satisfying the maximum input limitation of the VAE, thus addressing the difficulty of the original (Non PCA-based) DDTD to be applied to 3D strongly nonlinear optimization problems.
Furthermore, we demonstrated that the proposed PCA-based DDTD achieves elite material distributions with superior performances compared to Non PCA-based DDTD.
By solving MVMS minimization problems with/without considering large deformation and 3D compliant mechanism design problem, we validated the effectiveness of the proposed PCA-based DDTD. \\

\noindent\textbf{Funding}:
The second author was supported by JSPS KAKENHI (Grant No. 23H03799).

\section*{Declarations}

\textbf{Conflict of interest}: The authors declare that they have no conflict of interest. \\

\noindent\textbf{Replication of results}: The necessary information for a replication of the results are presented in the manuscript.
Interested readers may contact the corresponding author for further details regarding the implementation.

\bibliography{sn-bibliography}



\end{document}